\documentclass[runningheads]{llncs}
\usepackage[T1]{fontenc}
\usepackage{amsfonts}
\usepackage{amsmath}
\usepackage{amssymb}
\usepackage{appendix}
\usepackage{cancel}
\usepackage{hyperref}

\usepackage{graphicx}
\graphicspath{{figures/}}
\usepackage{color}

\usepackage{pgfplots}
\usepackage{circuitikz}
\usepackage{tikz-cd}
\usetikzlibrary{intersections}
\usepackage{bm}
\usetikzlibrary{positioning}
\usepgfplotslibrary{fillbetween}
\pgfplotsset{compat=1.5}
\usetikzlibrary{pgfplots.groupplots}
\usetikzlibrary{shapes.geometric,backgrounds}

\ctikzset{bipoles/length=.6cm}

\usepackage{tikz}
\usepackage{xifthen}

\pgfdeclarelayer{bg}    % declare background layer
\pgfsetlayers{bg,main}  % set the order of the layers (main is the standard layer)
\definecolor{beige}{RGB}{245, 245, 220}

\definecolor{darkgrey}{RGB}{75, 75, 75}
\definecolor{lightgrey}{RGB}{250, 250, 250}

\usetikzlibrary{calc, arrows, fit, positioning, patterns, decorations.pathreplacing, shapes}
\tikzstyle{dash} = [dashed, -latex,>=latex]
\tikzstyle{line} = [draw, -latex,>=latex]
\tikzstyle{smallbox} = [draw, minimum size=5.0mm]
\tikzstyle{box} = [draw, minimum size=7.0mm]
\tikzstyle{bigbox} = [draw, minimum size=10.0mm]
\tikzstyle{rectangle} = [draw, minimum width=10.0mm, minimum height=20.0mm]
\tikzstyle{switch} = [trapezium, trapezium angle=120, draw, rotate=90,  inner ysep=5pt, outer sep=5pt,
minimum height=7mm, minimum width=7mm]
\tikzstyle{roundbox} = [draw, circle, inner sep=0pt, minimum size=3mm]
\tikzstyle{clamped} = [draw, fill=darkgrey, minimum size=0.15cm]
\tikzstyle{msgcircle} = [shape=circle, draw, inner sep=0pt, minimum size=4mm, fill=white, font=\scriptsize]
\tikzstyle{darkmsgcircle} = [shape=circle, draw, inner sep=0pt, minimum size=4mm, fill=darkgrey, text=white, font=\scriptsize]
\tikzstyle{redmsgcircle} = [shape=circle, draw=red, inner sep=0pt, minimum size=4mm, text=red, font=\scriptsize]
\tikzstyle{reddarkmsgcircle} = [shape=circle, draw=red, inner sep=0pt, minimum size=4mm, fill=red, text=white, font=\scriptsize]
\tikzstyle{msgdoublecircle} = [shape=circle, double, double distance=1.5pt, draw, inner sep=0pt, minimum size=5mm, fill=white]
\tikzstyle{darkmsgdoublecircle} = [shape=circle, double, double distance=1.5pt, draw, inner sep=0pt, minimum size=5mm, fill=darkgrey, text=white, font=\bfseries]
\newcommand{\msg}[6]{
      \ifthenelse{\isin{#1}{left} \AND \isin{#2}{down}}{
            \coordinate (anchor) at ($({#3})!{#5}!({#4})$);
            \node[xshift=-6.0mm] at (anchor) {#6};
            \node[xshift=-1.0mm] at (anchor) {$\downarrow$};
      }{}
      \ifthenelse{\isin{#1}{right} \AND \isin{#2}{down}}{
            \coordinate (anchor) at ($({#3})!{#5}!({#4})$);
            \node[xshift=6.0mm] at (anchor) {#6};
            \node[xshift=1.0mm] at (anchor) {$\downarrow$};
      }{}

      \ifthenelse{\isin{#1}{down} \AND \isin{#2}{right}}{
            \coordinate (anchor) at ($({#3})!{#5}!({#4})$);
            \node[ yshift=-4.0mm] at (anchor) {#6};
            \node[yshift=-1.0mm] at (anchor) {$\rightarrow$};
      }{}
      \ifthenelse{\isin{#1}{up} \AND \isin{#2}{right}}{
            \coordinate (anchor) at ($({#3})!{#5}!({#4})$);
            \node[ yshift=4.0mm] at (anchor) {#6};
            \node[yshift=1.0mm] at (anchor) {$\rightarrow$};
      }{}

      \ifthenelse{\isin{#1}{down} \AND \isin{#2}{left}}{
            \coordinate (anchor) at ($({#3})!{#5}!({#4})$);
            \node[ yshift=-4.0mm] at (anchor) {#6};
            \node[yshift=-1.0mm] at (anchor) {$\leftarrow$};
      }{}
      \ifthenelse{\isin{#1}{up} \AND \isin{#2}{left}}{
            \coordinate (anchor) at ($({#3})!{#5}!({#4})$);
            \node[ yshift=4.0mm] at (anchor) {#6};
            \node[yshift=1.0mm] at (anchor) {$\leftarrow$};
      }{}

      \ifthenelse{\isin{#1}{left} \AND \isin{#2}{up}}{
            \coordinate (anchor) at ($({#3})!{#5}!({#4})$);
            \node[ xshift=-6.0mm] at (anchor) {#6};
            \node[xshift=-1.0mm] at (anchor) {$\uparrow$};
      }{}
      \ifthenelse{\isin{#1}{right} \AND \isin{#2}{up}}{
            \coordinate (anchor) at ($({#3})!{#5}!({#4})$);
            \node[ xshift=6.0mm] at (anchor) {#6};
            \node[xshift=1.0mm] at (anchor) {$\uparrow$};
      }{}
}

\newcommand{\msgcircle}[6]{
      \ifthenelse{\isin{#1}{left} \AND \isin{#2}{down}}{
            \coordinate (anchor) at ($({#3})!{#5}!({#4})$);
            \node[msgcircle,xshift=-5.0mm] at (anchor) {#6};
            \node[xshift=-1.5mm] at (anchor) {$\downarrow$};
      }{}
      \ifthenelse{\isin{#1}{right} \AND \isin{#2}{down}}{
            \coordinate (anchor) at ($({#3})!{#5}!({#4})$);
            \node[msgcircle,xshift=5.0mm] at (anchor) {#6};
            \node[xshift=1.5mm] at (anchor) {$\downarrow$};
      }{}

      \ifthenelse{\isin{#1}{down} \AND \isin{#2}{right}}{
            \coordinate (anchor) at ($({#3})!{#5}!({#4})$);
            \node[msgcircle, yshift=-5.0mm] at (anchor) {#6};
            \node[yshift=-2.0mm] at (anchor) {$\rightarrow$};
      }{}
      \ifthenelse{\isin{#1}{up} \AND \isin{#2}{right}}{
            \coordinate (anchor) at ($({#3})!{#5}!({#4})$);
            \node[msgcircle, yshift=5.0mm] at (anchor) {#6};
            \node[yshift=2.0mm] at (anchor) {$\rightarrow$};
      }{}

      \ifthenelse{\isin{#1}{down} \AND \isin{#2}{left}}{
            \coordinate (anchor) at ($({#3})!{#5}!({#4})$);
            \node[msgcircle, yshift=-5.0mm] at (anchor) {#6};
            \node[yshift=-2.0mm] at (anchor) {$\leftarrow$};
      }{}
      \ifthenelse{\isin{#1}{up} \AND \isin{#2}{left}}{
            \coordinate (anchor) at ($({#3})!{#5}!({#4})$);
            \node[msgcircle, yshift=5.0mm] at (anchor) {#6};
            \node[yshift=2.0mm] at (anchor) {$\leftarrow$};
      }{}

      \ifthenelse{\isin{#1}{left} \AND \isin{#2}{up}}{
            \coordinate (anchor) at ($({#3})!{#5}!({#4})$);
            \node[msgcircle, xshift=-5.0mm] at (anchor) {#6};
            \node[xshift=-1.5mm] at (anchor) {$\uparrow$};
      }{}
      \ifthenelse{\isin{#1}{right} \AND \isin{#2}{up}}{
            \coordinate (anchor) at ($({#3})!{#5}!({#4})$);
            \node[msgcircle, xshift=5.0mm] at (anchor) {#6};
            \node[xshift=1.5mm] at (anchor) {$\uparrow$};
      }{}
}

\newcommand{\darkmsg}[6]{
      \ifthenelse{\isin{#1}{left} \AND \isin{#2}{down}}{
            \coordinate (anchor) at ($({#3})!{#5}!({#4})$);
            \node[darkmsgcircle, xshift=-5mm] at (anchor) {#6};
            \node[xshift=-1.5mm] at (anchor) {$\downarrow$};
      }{}
      \ifthenelse{\isin{#1}{right} \AND \isin{#2}{down}}{
            \coordinate (anchor) at ($({#3})!{#5}!({#4})$);
            \node[darkmsgcircle, xshift=5mm] at (anchor) {#6};
            \node[xshift=1.5mm] at (anchor) {$\downarrow$};
      }{}

      \ifthenelse{\isin{#1}{down} \AND \isin{#2}{right}}{
            \coordinate (anchor) at ($({#3})!{#5}!({#4})$);
            \node[darkmsgcircle, yshift=-5.0mm] at (anchor) {#6};
            \node[yshift=-2.0mm] at (anchor) {$\rightarrow$};
      }{}
      \ifthenelse{\isin{#1}{up} \AND \isin{#2}{right}}{
            \coordinate (anchor) at ($({#3})!{#5}!({#4})$);
            \node[darkmsgcircle, yshift=5.0mm] at (anchor) {#6};
            \node[yshift=2.0mm] at (anchor) {$\rightarrow$};
      }{}

      \ifthenelse{\isin{#1}{down} \AND \isin{#2}{left}}{
            \coordinate (anchor) at ($({#3})!{#5}!({#4})$);
            \node[darkmsgcircle, yshift=-5.0mm] at (anchor) {#6};
            \node[yshift=-2.0mm] at (anchor) {$\leftarrow$};
      }{}
      \ifthenelse{\isin{#1}{up} \AND \isin{#2}{left}}{
            \coordinate (anchor) at ($({#3})!{#5}!({#4})$);
            \node[darkmsgcircle, yshift=5.0mm] at (anchor) {#6};
            \node[yshift=2.0mm] at (anchor) {$\leftarrow$};
      }{}

      \ifthenelse{\isin{#1}{left} \AND \isin{#2}{up}}{
            \coordinate (anchor) at ($({#3})!{#5}!({#4})$);
            \node[darkmsgcircle, xshift=-5.0mm] at (anchor) {#6};
            \node[xshift=-1.5mm] at (anchor) {$\uparrow$};
      }{}
      \ifthenelse{\isin{#1}{right} \AND \isin{#2}{up}}{
            \coordinate (anchor) at ($({#3})!{#5}!({#4})$);
            \node[darkmsgcircle, xshift=5.0mm] at (anchor) {#6};
            \node[xshift=1.5mm] at (anchor) {$\uparrow$};
      }{}
}

\newcommand{\redbackmsg}[6]{
      \ifthenelse{\isin{#1}{left} \AND \isin{#2}{down}}{
            \coordinate (anchor) at ($({#3})!{#5}!({#4})$);
            \node[reddarkmsgcircle, xshift=-5mm] at (anchor) {#6};
            \node[xshift=-1.5mm] at (anchor) {$\downarrow$};
      }{}
      \ifthenelse{\isin{#1}{right} \AND \isin{#2}{down}}{
            \coordinate (anchor) at ($({#3})!{#5}!({#4})$);
            \node[reddarkmsgcircle, xshift=5mm] at (anchor) {#6};
            \node[xshift=1.5mm] at (anchor) {$\downarrow$};
      }{}

      \ifthenelse{\isin{#1}{down} \AND \isin{#2}{right}}{
            \coordinate (anchor) at ($({#3})!{#5}!({#4})$);
            \node[reddarkmsgcircle, yshift=-5.0mm] at (anchor) {#6};
            \node[yshift=-2.0mm] at (anchor) {$\rightarrow$};
      }{}
      \ifthenelse{\isin{#1}{up} \AND \isin{#2}{right}}{
            \coordinate (anchor) at ($({#3})!{#5}!({#4})$);
            \node[reddarkmsgcircle, yshift=5.0mm] at (anchor) {#6};
            \node[yshift=2.0mm] at (anchor) {$\rightarrow$};
      }{}

      \ifthenelse{\isin{#1}{down} \AND \isin{#2}{left}}{
            \coordinate (anchor) at ($({#3})!{#5}!({#4})$);
            \node[reddarkmsgcircle, yshift=-5.0mm] at (anchor) {#6};
            \node[yshift=-2.0mm] at (anchor) {$\leftarrow$};
      }{}
      \ifthenelse{\isin{#1}{up} \AND \isin{#2}{left}}{
            \coordinate (anchor) at ($({#3})!{#5}!({#4})$);
            \node[reddarkmsgcircle, yshift=5.0mm] at (anchor) {#6};
            \node[yshift=2.0mm] at (anchor) {$\leftarrow$};
      }{}

      \ifthenelse{\isin{#1}{left} \AND \isin{#2}{up}}{
            \coordinate (anchor) at ($({#3})!{#5}!({#4})$);
            \node[reddarkmsgcircle, xshift=-5.0mm] at (anchor) {#6};
            \node[xshift=-1.5mm] at (anchor) {$\uparrow$};
      }{}
      \ifthenelse{\isin{#1}{right} \AND \isin{#2}{up}}{
            \coordinate (anchor) at ($({#3})!{#5}!({#4})$);
            \node[reddarkmsgcircle, xshift=5.0mm] at (anchor) {#6};
            \node[xshift=1.5mm] at (anchor) {$\uparrow$};
      }{}
}

\newcommand{\redmsg}[6]{
      \ifthenelse{\isin{#1}{left} \AND \isin{#2}{down}}{
            \coordinate (anchor) at ($({#3})!{#5}!({#4})$);
            \node[redmsgcircle, xshift=-5mm] at (anchor) {#6};
            \node[xshift=-1.5mm] at (anchor) {$\downarrow$};
      }{}
      \ifthenelse{\isin{#1}{right} \AND \isin{#2}{down}}{
            \coordinate (anchor) at ($({#3})!{#5}!({#4})$);
            \node[redmsgcircle, xshift=5mm] at (anchor) {#6};
            \node[xshift=1.5mm] at (anchor) {$\downarrow$};
      }{}

      \ifthenelse{\isin{#1}{down} \AND \isin{#2}{right}}{
            \coordinate (anchor) at ($({#3})!{#5}!({#4})$);
            \node[redmsgcircle, yshift=-5.0mm] at (anchor) {#6};
            \node[yshift=-2.0mm] at (anchor) {$\rightarrow$};
      }{}
      \ifthenelse{\isin{#1}{up} \AND \isin{#2}{right}}{
            \coordinate (anchor) at ($({#3})!{#5}!({#4})$);
            \node[redmsgcircle, yshift=5.0mm] at (anchor) {#6};
            \node[yshift=2.0mm] at (anchor) {$\rightarrow$};
      }{}

      \ifthenelse{\isin{#1}{down} \AND \isin{#2}{left}}{
            \coordinate (anchor) at ($({#3})!{#5}!({#4})$);
            \node[redmsgcircle, yshift=-5.0mm] at (anchor) {#6};
            \node[yshift=-2.0mm] at (anchor) {$\leftarrow$};
      }{}
      \ifthenelse{\isin{#1}{up} \AND \isin{#2}{left}}{
            \coordinate (anchor) at ($({#3})!{#5}!({#4})$);
            \node[redmsgcircle, yshift=5.0mm] at (anchor) {#6};
            \node[yshift=2.0mm] at (anchor) {$\leftarrow$};
      }{}

      \ifthenelse{\isin{#1}{left} \AND \isin{#2}{up}}{
            \coordinate (anchor) at ($({#3})!{#5}!({#4})$);
            \node[redmsgcircle, xshift=-5.0mm] at (anchor) {#6};
            \node[xshift=-1.5mm] at (anchor) {$\uparrow$};
      }{}
      \ifthenelse{\isin{#1}{right} \AND \isin{#2}{up}}{
            \coordinate (anchor) at ($({#3})!{#5}!({#4})$);
            \node[redmsgcircle, xshift=5.0mm] at (anchor) {#6};
            \node[xshift=1.5mm] at (anchor) {$\uparrow$};
      }{}
}

\newcommand{\bwmsg}[6]{
      \ifthenelse{\isin{#1}{left} \AND \isin{#2}{down}}{
            \coordinate (anchor) at ($({#3})!{#5}!({#4})$);
            \node[msgdoublecircle, xshift=-5.5mm] at (anchor) {#6};
            \node[xshift=-1.5mm] at (anchor) {$\downarrow$};
      }{}
      \ifthenelse{\isin{#1}{right} \AND \isin{#2}{down}}{
            \coordinate (anchor) at ($({#3})!{#5}!({#4})$);
            \node[msgdoublecircle, xshift=5.5mm] at (anchor) {#6};
            \node[xshift=1.5mm] at (anchor) {$\downarrow$};
      }{}

      \ifthenelse{\isin{#1}{down} \AND \isin{#2}{right}}{
            \coordinate (anchor) at ($({#3})!{#5}!({#4})$);
            \node[msgdoublecircle, yshift=-6.0mm] at (anchor) {#6};
            \node[yshift=-2.0mm] at (anchor) {$\rightarrow$};
      }{}
      \ifthenelse{\isin{#1}{up} \AND \isin{#2}{right}}{
            \coordinate (anchor) at ($({#3})!{#5}!({#4})$);
            \node[msgdoublecircle, yshift=6.0mm] at (anchor) {#6};
            \node[yshift=2.0mm] at (anchor) {$\rightarrow$};
      }{}

      \ifthenelse{\isin{#1}{down} \AND \isin{#2}{left}}{
            \coordinate (anchor) at ($({#3})!{#5}!({#4})$);
            \node[msgdoublecircle, yshift=-6.0mm] at (anchor) {#6};
            \node[yshift=-2.0mm] at (anchor) {$\leftarrow$};
      }{}
      \ifthenelse{\isin{#1}{up} \AND \isin{#2}{left}}{
            \coordinate (anchor) at ($({#3})!{#5}!({#4})$);
            \node[msgdoublecircle, yshift=6.0mm] at (anchor) {#6};
            \node[yshift=2.0mm] at (anchor) {$\leftarrow$};
      }{}

      \ifthenelse{\isin{#1}{left} \AND \isin{#2}{up}}{
            \coordinate (anchor) at ($({#3})!{#5}!({#4})$);
            \node[msgdoublecircle, xshift=-5.5mm] at (anchor) {#6};
            \node[xshift=-1.5mm] at (anchor) {$\uparrow$};
      }{}
      \ifthenelse{\isin{#1}{right} \AND \isin{#2}{up}}{
            \coordinate (anchor) at ($({#3})!{#5}!({#4})$);
            \node[msgdoublecircle, xshift=5.5mm] at (anchor) {#6};
            \node[xshift=1.5mm] at (anchor) {$\uparrow$};
      }{}
}

\newcommand{\bwdarkmsg}[6]{
      \ifthenelse{\isin{#1}{left} \AND \isin{#2}{down}}{
            \coordinate (anchor) at ($({#3})!{#5}!({#4})$);
            \node[darkmsgdoublecircle, xshift=-5.5mm] at (anchor) {#6};
            \node[xshift=-1.5mm] at (anchor) {$\downarrow$};
      }{}
      \ifthenelse{\isin{#1}{right} \AND \isin{#2}{down}}{
            \coordinate (anchor) at ($({#3})!{#5}!({#4})$);
            \node[darkmsgdoublecircle, xshift=5.5mm] at (anchor) {#6};
            \node[xshift=1.5mm] at (anchor) {$\downarrow$};
      }{}

      \ifthenelse{\isin{#1}{down} \AND \isin{#2}{right}}{
            \coordinate (anchor) at ($({#3})!{#5}!({#4})$);
            \node[darkmsgdoublecircle, yshift=-6.0mm] at (anchor) {#6};
            \node[yshift=-2.0mm] at (anchor) {$\rightarrow$};
      }{}
      \ifthenelse{\isin{#1}{up} \AND \isin{#2}{right}}{
            \coordinate (anchor) at ($({#3})!{#5}!({#4})$);
            \node[darkmsgdoublecircle, yshift=6.0mm] at (anchor) {#6};
            \node[yshift=2.0mm] at (anchor) {$\rightarrow$};
      }{}

      \ifthenelse{\isin{#1}{down} \AND \isin{#2}{left}}{
            \coordinate (anchor) at ($({#3})!{#5}!({#4})$);
            \node[darkmsgdoublecircle, yshift=-6.0mm] at (anchor) {#6};
            \node[yshift=-2.0mm] at (anchor) {$\leftarrow$};
      }{}
      \ifthenelse{\isin{#1}{up} \AND \isin{#2}{left}}{
            \coordinate (anchor) at ($({#3})!{#5}!({#4})$);
            \node[darkmsgdoublecircle, yshift=6.0mm] at (anchor) {#6};
            \node[yshift=2.0mm] at (anchor) {$\leftarrow$};
      }{}

      \ifthenelse{\isin{#1}{left} \AND \isin{#2}{up}}{
            \coordinate (anchor) at ($({#3})!{#5}!({#4})$);
            \node[darkmsgdoublecircle, xshift=-5.5mm] at (anchor) {#6};
            \node[xshift=-1.5mm] at (anchor) {$\uparrow$};
      }{}
      \ifthenelse{\isin{#1}{right} \AND \isin{#2}{up}}{
            \coordinate (anchor) at ($({#3})!{#5}!({#4})$);
            \node[darkmsgdoublecircle, xshift=5.5mm] at (anchor) {#6};
            \node[xshift=1.5mm] at (anchor) {$\uparrow$};
      }{}
}

\tikzset{mainstyle/.style={fill=white, draw=black, shape=rectangle, align=center}}

\tikzset{dstyle/.style={mainstyle, minimum size=4mm, inner sep=0pt, text width=4mm}}

\tikzset{sstyle/.style={mainstyle, minimum size=5mm, inner sep=0pt, text width=5mm}}

\tikzset{ostyle/.style={fill=darkgrey, draw=black, shape=rectangle, minimum size=0.2cm, inner sep=0pt, text width=2mm}}

\tikzstyle{observation}=[ostyle]
\tikzstyle{deterministic}=[dstyle]
\tikzstyle{stochastic}=[sstyle]

\tikzstyle{filter}=[mainstyle, minimum width=1cm, minimum height=0.5cm]
\tikzstyle{selector}=[fill=white, draw=black, shape=trapezium, rotate=180, minimum width=1cm, minimum height=0.5cm]

\newcommand*\wcircled[1]{\tikz[baseline=(char.base)]{\node[shape=circle,draw,minimum size=4mm,inner sep=0pt] (char) {#1};}}

\def\cN{\mathcal{N}}

\def\-{\text{-}}
\def\+{\text{+}}

\def\tr{\mathrm{Tr}}
\newcommand\given[1][]{\:#1\vert\:}

\begin{document}
\title{Message passing-based inference in an autoregressive active inference agent}
\author{Wouter M. Kouw\inst{1}\and
Tim N. Nisslbeck\inst{1}\and
Wouter L.N. Nuijten\inst{1,2}}
% %
\authorrunning{W.M. Kouw et al.}
\institute{\textsuperscript{1}Eindhoven University of Technology, Eindhoven, Netherlands\\
\textsuperscript{2}Lazy Dynamics B.V., Eindhoven, Netherlands\\
\email{w.m.kouw@tue.nl}
}
\maketitle              
\begin{abstract}
We present the design of an autoregressive active inference agent in the form of message passing on a factor graph. Expected free energy is derived and distributed across a planning graph. The proposed agent is validated on a robot navigation task, demonstrating exploration and exploitation in a continuous-valued observation space with bounded continuous-valued actions. Compared to a classical optimal controller, the agent modulates action based on predictive uncertainty, arriving later but with a better model of the robot's dynamics.
\keywords{Intelligent agents \and Free energy minimization \and Active inference \and Autoregressive models \and Factor graphs \and Message passing}
\end{abstract}
\section{Introduction}
Active inference is a comprehensive framework that unifies perception, planning, and learning under the free energy principle, offering a promising approach to designing autonomous agents \cite{friston2006free,parr2022active}. We present the design of an active inference agent implemented as a message passing procedure on a Forney-style factor graph \cite{van2019simulating,van2022active}. The agent is built on an autoregressive model, making continuous-valued observations and inferring bounded continuous-valued actions \cite{kouw2023information,nisslbeck2025marx}. We show that leveraging the factor graph approach produces a distributed, efficient and modular implementation \cite{de2017factor,friston2017graphical,parr2019neuronal,de2025expected}. 

Probabilistic graphical models have long been a unifying framework for the design and analysis of information processing systems, including signal processing, optimal controllers, and artificially intelligent agents \cite{hewitt1977viewing,loeliger2007factor,hoffmann2017linear,palmieri2021unified,van2022active}. Many famous algorithms can be written as message passing algorithms, including Kalman filtering, model-predictive control, and dynamic programming \cite{loeliger2007factor,palmieri2021unified}. 
However, it can be a challenge to formulate new algorithms due to the requirement of local access to variables and the difficulty of deriving backwards messages. We highlight some of these challenges, and contribute with
\begin{itemize}
    \item[$\bullet$] the derivation of expected free energy minimization in a multivariate autoregressive model with continuous-valued observations and bounded continuous-valued actions (Sec.~\ref{sec:fe-actions}), and
    \item[$\bullet$] the formulation of the planning model as a factor graph with marginal distribution updates based on messages passed along the graph (Figure \ref{fig:ffg-planning-Tstep}).
\end{itemize}
We validate the proposed design on a robot navigation task, comparing the agent to an adaptive model-predictive controller.

\section{Problem statement}
We focus on the class of discrete-time stochastic nonlinear dynamical systems with state $z_k \in \mathbb{R}^{D_z}$, control $u_k \in \mathbb{R}^{D_u}$, and observation $y_k \in \mathbb{R}^{D_y}$ at time $k$. Their evolution is governed by a state transition function $f$ and an observation function $g$:
\begin{align} \label{eq:problem}
    z_k = f(z_{k-1}, u_k) + w_k \, , \qquad
    y_k = g(z_k) + v_k \, ,
\end{align}
where $w_k, v_k$ are stochastic contributions. The agent only receives noisy outputs $y_k \in \mathbb{R}^{D_y}$ from a system and sends control inputs $u_k \in \mathbb{U} \subset \mathbb{R}^{D_u}$ back. It must drive the system to output $y_{*}$ without knowledge of the system's dynamics. Performance is measured with free energy (which in the proposed model is equal to the negative log evidence), Euclidean distance to goal, and the 2-norm magnitude of controls, over the course of a trial of length $T$.

\section{Model specification}
The model is autoregressive in nature, meaning that the system output at time $k$ is predicted from the system input $u_k$, $M_u$ previous system inputs $\bar{u}_k$ and $M_y$ previous system outputs $\bar{y}_k$:
\begin{align}
    \bar{u}_k = \begin{bmatrix} u_{k-1} \\ \vdots \\ u_{k-{M_u}} \end{bmatrix} \, , \ \ \bar{y}_k = \begin{bmatrix} y_{k-1} \\ \vdots \\ y_{k-{M_y}} \end{bmatrix} \, , \ \ x_k = \begin{bmatrix} u_k \\ \bar{u}_k \\ \bar{y}_k \end{bmatrix} . 
\end{align}
The vector $x_k$ is the concatenation of these elements and has dimension $D_x = D_u (M_u + 1) + D_y M_y$. 
Our likelihood function is based on a Gaussian distribution 
\begin{align} \label{eq:likelihood}
    p(y_k \given \Theta, u_k, \bar{u}_k, \bar{y}_k) = \mathcal{N}\big(y_k \given A^{\intercal} x_k, W^{-1} \big) \, ,
\end{align}
where $A \in \mathbb{R}^{D_x \times D_y}$ is a regression coefficient matrix and $W \in \mathbb{R}_{+}^{\, D_y \times D_y}$ is a precision matrix. Let $\Theta = (A,W)$ refer to the parameters jointly.
Their prior distribution is a matrix normal Wishart distribution \cite[D175]{Soch}:
\begin{align} \label{eq:param-priors}
    p(\Theta) &= \mathcal{MNW}\big(A, W \given M_0, \Lambda_0^{-1}, \Omega_0^{-1}, \nu_0 \big) \\
    &=  \mathcal{MN}\big(A \given M_0, \Lambda_0^{-1}, W^{-1} \big) \, \mathcal{W}\big(W \given \Omega_0^{-1}, \nu_0 \big) .
\end{align}
The prior distributions over control inputs are independent Gaussian distributions, as are the goal prior distributions for future observations:
\begin{align}
    p(u_k) = \cN(u_k \given 0, \Upsilon^{-1}) \, , \qquad p(y_t \given y_{*}) = \mathcal{N}(y_t \given m_{*}, S_{*}) \label{eq:control-goal-prior} \, ,
\end{align}
where $\Upsilon$ is a precision matrix and $y_{*} = (m_{*},S_{*})$ are the goal mean vector and covariance matrix. 

\section{Inference} 

\subsection{Learning} \label{sec:fe-params}
We use Bayesian filtering to update parameter beliefs given $y_k, u_k$ \cite{sarkka2013bayesian,nisslbeck2025marx}:
\begin{align} \label{eq:bayesian-filtering}
    \underbrace{p\big(\Theta \given \mathcal{D}_{k} \big)}_{\text{posterior}}  =  \frac{\overbrace{p\big(y_k \given \Theta, u_k, \bar{u}_k, \bar{y}_k \big)}^{\text{likelihood}}}{\underbrace{p\big(y_{k} \given u_{k}, \mathcal{D}_{k\-1}\big)}_{\text{evidence}}} \, \underbrace{p\big(\Theta \given \mathcal{D}_{k\-1}\big)}_{\text{prior}} . 
\end{align}
where $\mathcal{D}_k = \{y_i, u_i\}_{i=1}^{k}$ is short-hand for data up to time $k$. Note that the memories $\bar{u}_k, \bar{y}_k$ are subsets of $\mathcal{D}_{k-1}$. The evidence term is the evaluation of the observation $y_k$ under the predictive distribution, obtained by marginalizing the likelihood over the parameters \cite{nisslbeck2025factor}.

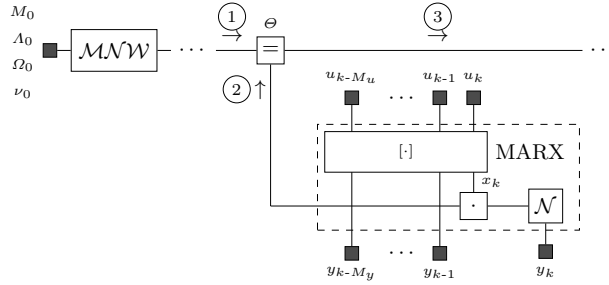
\begin{figure}[bth]
    \centering
    \vspace{-10pt}
    \scalebox{0.9}{\begin{tikzpicture}

    % likelihood
    \node [style=deterministic, minimum width=24mm, minimum height=6mm, align=center] (cat) {$\scriptstyle{[ \cdot ]}$};
    \node [style=deterministic] (dot) at (1.,-.8) {$\cdot$};
    \node [right=0mm of cat] (title) {MARX};
    \node [style=stochastic, right=6mm of dot] (N) {$\mathcal{N}$};
    \draw [-] (dot) -- (N);
    \draw [-] (dot) -- node[right] {$\scriptstyle{x_k}$} ++(0.,.5);
    
    \node [style=observation] (u_k_node) at (1.0,0.8) {};
    \node [style=observation] (u_kmin1_node) at (0.5,0.8) {};
    \node [left=1mm of u_kmin1_node] (u_kmin_dots) {$\cdots$};
    \node [style=observation] (u_kminM_node) at (-0.8,0.8) {};
    \node [above of=u_k_node,     node distance=3mm] (u_k) {$\scriptstyle{u_{k}}$};
    \node [above of=u_kmin1_node, node distance=3mm] (u_kmin1) {$\scriptstyle{u_{k\text{-}1}}$};
    \node [above of=u_kminM_node, node distance=3mm] (u_kminM) {$\scriptstyle{u_{k\text{-}M_u}}$};
    \draw[-] (u_k_node)     -- ++(0.0, -0.5);
    \draw[-] (u_kmin1_node) -- ++(0.0, -0.5);
    \draw[-] (u_kminM_node) -- ++(0.0, -0.5);
    
    \node [style=observation] (y_kmin1_node) at (0.5,-1.5) {};
    \node [left=1mm of y_kmin1_node] (y_kmin_dots) {$\cdots$};
    \node [style=observation] (y_kminM_node) at (-0.8,-1.5) {};
    \node [below of=y_kmin1_node, node distance=3mm] (y_kmin1) {$\scriptstyle{y_{k\text{-}1}}$};
    \node [below of=y_kminM_node, node distance=3mm] (y_kminM) {$\scriptstyle{y_{k\text{-}M_y}}$};
    \draw[-] (y_kmin1_node) -- ++(0.0, 1.2);
    \draw[-] (y_kminM_node) -- ++(0.0, 1.2);
    
    \node [style=observation, below = 3.1mm of N] (y_k_node) {};
    \node [below of=y_k_node, node distance=3mm] (y_k) {$\scriptstyle{y_k}$};
    \draw [-] (N) -- (y_k_node);
    \node[dashed, fit=(cat)(title)(N), draw, inner sep=1mm] (box) {};
    
    % parameter chain
    \node [style=deterministic] (eqN) at (-2.,1.5) {$=$};
    \node [] (eqN_) at (-2.,-.8) {};
    \draw [-] (eqN) -- (eqN_.center);
    \draw [-] (eqN_.center) -- (dot);
    
    % Prior
    \node [left =6mm of eqN] (prevzeta) {$\cdots$};
    \draw [-] (prevzeta) -- (eqN);
    \node [shape=rectangle, draw=black, inner ysep=2mm, minimum width=10mm, left=2mm of prevzeta] (zetaprior) {$\mathcal{MNW}$};
    \draw [-] (zetaprior) -- (prevzeta);
    \node [style=observation, left = 2mm of zetaprior] (zetaprior_params) {};
    \node [left of=zetaprior_params, node distance=4mm] (zetaprior_params0) {$\begin{matrix} \scriptstyle{M_0} \\ \scriptstyle{\Lambda_0} \\ \scriptstyle{\Omega_0} \\ \scriptstyle{\nu_0} \end{matrix}$};
    \draw [-] (zetaprior_params) -- (zetaprior);

    % Posterior
    \node [right = 44mm of eqN] (zetapost) {$\cdots$};
    \draw [-] (eqN) -- (zetapost);
    \node [above of=eqN, node distance=4mm] {$\scriptstyle{\Theta}$};

    % messages
    \msgcircle{up}{right}{prevzeta}{eqN}{0.5}{1};
    \msgcircle{left}{up}{eqN}{eqN_}{0.25}{2};
    \msgcircle{up}{right}{eqN}{zetapost}{0.5}{3};
    % \msgcircle{left}{down}{N}{y}{0.7}{4};
    
\end{tikzpicture}}
    \caption{Forney-style factor graph of one time step (separated by dots) of Bayesian filtering. Edges represent random variables and nodes operations on those variables. Black squares represent observed variables or set parameters, and the dotted box represents a custom node, composed of the nodes within. Message $1$ is the prior belief over parameters and message $2$ the likelihood-based update. These are are multiplied at the equality node, yielding the marginal posterior distribution (message $3$).} \vspace{-10pt}
    \label{fig:ffg-paramest}
\end{figure}

\noindent We express the Bayesian filtering procedure as message passing on the factor graph\footnote{For excellent introductions to the factor graph approach, see \cite{loeliger2004introduction,csenoz2021variational}.} shown in Figure \ref{fig:ffg-paramest}. Message $\wcircled{1}$ is the prior distribution on $\Theta$, 
\begin{align}
    \wcircled{1} = \mathcal{MNW}(A,W \given M_{k\-1}, \Lambda_{k\-1}, \Omega_{k\-1}, \nu_{k\-1}) \, .
\end{align} 
Message $\wcircled{2}$ originates from the MARX likelihood function and is an improper matrix normal Wishart distribution \cite[Lemma~2]{nisslbeck2025marx},
\begin{align} \label{eq:message2}
    \wcircled{2} = \mathcal{MNW}(A,W \given \bar{M}_k, \bar{\Lambda}_k^{-1}, \bar{\Omega}^{-1}_k, \bar{\nu}_k) \, ,
\end{align}
with parameters based on data and buffers at time $k$,
\begin{align}
    \bar{\nu}_k = 2 \! - \! D_x \! + \! D_y \, , \
    \bar{\Lambda}_k = x_kx_k^{\intercal} \, , \
    \bar{M}_k = (x_kx_k^{\intercal})^{-1} x_ky_k^{\intercal} \, , \ 
    \bar{\Omega}_k = 0_{D_y \times D_y} \, .
\end{align}

\noindent It is improper because $\bar{\omega}_k$ is singular. But when multiplied with the prior distribution, it produces the conjugate posterior distribution exactly \cite[Thm.~1]{nisslbeck2025marx}.
This multiplication occurs in the equality node and produces message $\wcircled{3}$:
\begin{align} \label{eq:truepost}
    \wcircled{3} = p(\Theta \given \mathcal{D}_k)  =  \mathcal{MNW}(A, W \given M_k, \Lambda^{-1}_k, \Omega^{-1}_k, \nu_k) \, .
\end{align}
The parameters of this distribution are
\begin{align} \label{eq:postparams}
\nu_k
&= \nu_{k\-1} + 1 , \\
\Lambda_k
&= \Lambda_{k\-1} + x_k x_k^\intercal , \\
M_k
&= (\Lambda_{k\-1} + x_k x_k^\intercal)^{-1}(\Lambda_{k\-1}M_{k\-1} + x_k y_k^\intercal)  , \\
\Omega_k 
&= \Omega_{k\-1} + y_k y_k^\intercal  +  M_{k\-1}^{\intercal}\Lambda_{k\-1}M_{k\-1} - \\
& \qquad \qquad (\Lambda_{k\-1}M_{k\-1} \! + \! x_k y_k^\intercal)^{\intercal} (\Lambda_{k\-1} \! + \! x_k x_k^\intercal)^{-1} (\Lambda_{k\-1} M_{k\-1} \! + \! x_k y_k^\intercal) \,  . \nonumber
\end{align}

Marginalizing the Gaussian likelihood in Eq.~\ref{eq:likelihood} over the parameter posterior distribution (Eq.~\ref{eq:truepost}) yields a multivariate location-scale T-distribution \cite{nisslbeck2025factor}:
    \begin{align} 
        p(y_k | u_k, \mathcal{D}_{k}) \!
          = \! \int \! p(y_k | \Theta, u_k, \bar{u}_k, \bar{y}_k) p(\Theta | \mathcal{D}_{k})  \mathrm{d}\Theta 
         = \! \mathcal{T}_{\eta_k} \big(y_k | \mu_k(u_k), \Sigma_k(u_k) \big) , \label{eq:postpred}
    \end{align}
with $\eta_k = \nu_{k} - D_y + 1$ degrees of freedom and a mean and covariance of
\begin{align}
    \mu_k(u_k) = M_{k}^{\intercal} \begin{bmatrix} u_k \\ \bar{u}_t \\ \bar{y}_t \end{bmatrix} , \
    \Sigma_k(u_k) = \frac{1}{\nu_{k} \! - \! D_y \! + \! 1} \Omega_{k} \Big(1 \! + \! \begin{bmatrix} u_t \\ \bar{u}_t \\ \bar{y}_t \end{bmatrix}^{\intercal} \Lambda_{k}^{-1} \begin{bmatrix} u_t \\ \bar{u}_t \\ \bar{y}_t \end{bmatrix} \Big) .
\end{align}
The subscripts under $\mu$ and $\Sigma$ indicate which parameters were used, i.e., here they refer to $M_k, \Lambda_k, \Omega_k$ and $\nu_k$.

\subsection{Actions} \label{sec:fe-actions}

\subsubsection{Planning}
We start by building a generative model for the input and output at time $t = k+1$:
\begin{align} \label{eq:p-future}
    p(y_t, \Theta, u_t \given \mathcal{D}_k) =   p(y_t \given \Theta, u_t, \bar{u}_t, \bar{y}_t) \, p(\Theta \given \mathcal{D}_k) \, p(u_t)  .
\end{align}
Note that $\bar{u}_t$ and $\bar{y}_t$ are absent on the left-hand side because, at time $t = k+1$, these buffers are subsets of $\mathcal{D}_k$. 
We want the agent to pursue a target, a specific future observation. To do so, we first isolate the marginal distribution $p(y_t)$, 
\begin{align} \label{eq:decomp-futurejoint}
    &p(y_t \given \Theta, u_t, \bar{u}_t, \bar{y}_t) p(\Theta \given \mathcal{D}_k)  =  p(\Theta \given y_t, u_t, \mathcal{D}_k) p(y_t) ,
\end{align}
and then constrain it to be the goal prior, $p(y_t) \rightarrow p(y_t \given y_{*})$. 
We use Bayes' rule in the reverse direction to relate the distribution over parameters given the future output and input, to known distributions:
\begin{align} \label{eq:futureparampost-bayes}
    p(\Theta \given y_t, u_t, \mathcal{D}_k) = \frac{p(y_t \given \Theta, u_t, \bar{u}_t, \bar{y}_t) p(\Theta \given \mathcal{D}_k)}{p(y_t \given u_t, \mathcal{D}_k)} \, .
\end{align}
To obtain an approximate marginal posterior distribution for the action $u_t$, we form an expected free energy functional,
\begin{align} \label{eq:EFE-planning}
\mathcal{F}_k[q] = \mathbb{E}_{q(y_t \given \Theta, u_t, \bar{u}_t, \bar{y}_t)} \Big[ \mathbb{E}_{q(\Theta, u_t)} \big[ \ln \frac{q(\Theta, u_t)}{p(y_t, \Theta, u_t, \given y_{*}, \mathcal{D}_k)} \big] \Big] \, .
\end{align}
The variational model is $q(y_t, \Theta, u_t, \bar{u}_t, \bar{y}_t) = q(y_t \given \Theta, u_t, \bar{u}_t, \bar{y}_t) q(\Theta) q(u_t)$. The likelihood and parameter factors are not free variational distributions but fixed to the same form as the likelihood and parameter factors of the generative model:
\begin{align} \label{eq:variational-model}
    q(y_t \given \Theta, u_t, \bar{u}_t, \bar{y}_t) &=  p(y_t \given \Theta, u_t, \bar{u}_t, \bar{y}_t) = \mathcal{N}\big(y_t \given A^{\intercal} x_t, W^{-1} \big) \\ 
    q(\Theta) &= p(\Theta \given \mathcal{D}_k) =  \mathcal{MNW}(A, W \given M_k, \Lambda^{-1}_k, \Omega^{-1}_k, \nu_k) \, .
\end{align}
We then minimize this expected free energy functional with respect to the variational distribution $q(u_t)$:
\begin{align} \label{eq:optimal-q}
    q^{*}(u_t) = \underset{q \, \in \, Q}{\arg \min} \ \mathcal{F}_k[q] \, .
\end{align}
where $Q$ represents the set of candidate variational distributions.
\begin{theorem} \label{eq:thm-qu}
    The optimal variational posterior $q^{*}(u_t)$ under the free energy functional defined in \eqref{eq:EFE-planning} is proportional to a prior times a likelihood,
    \begin{align}
        q^{*}(u_t) \propto p(u_t) \exp\big( - G(u_t) \big) \, ,
    \end{align}
    where $G$ is the sum of a mutual information and a cross-entropy term
    \begin{align}
        G(u_t) \! = \! -\mathbb{E}_{p(y_t, \Theta | u_t, \mathcal{D}_k)}\big[ \! \ln \! \frac{p(y_t, \Theta \given u_t,  \mathcal{D}_k)}{p(y_t | u_t,  \mathcal{D}_k) p(\Theta | \mathcal{D}_k)} \big] \! - \! \mathbb{E}_{p(y_t | u_t,  \mathcal{D}_k)}\big[ \ln p(y_t | y_{*} ) \big]  . \label{eq:EFEfunction}
    \end{align}
\end{theorem}
\noindent The proof can be found in Appendix \ref{proof:thm1}. 
\begin{corollary} \label{cor1}
    The expected free energy function $G(u_t)$ evaluates to: \begin{align} 
        G(u_t) = \mathrm{constants} -  \frac{1}{2} \ln | \Sigma_t(u_t) | +  \frac{1}{2} \tr\Big[S_{*}^{-1}  \big( \Sigma_t(u_t) \frac{\eta_t}{\eta_t \! - \! 2}  +  \Xi(u_t) \big) \Big] \, . \label{eq:expJu}
    \end{align}
    where  $\Xi(u_t) = (\mu_t(u_t)  -   m_{*} )(\mu_t(u_t) \! - \! m_{*} )^{\intercal}$.
\end{corollary}
\noindent The proof is also in Appendix \ref{proof:cor1}.

Figure \ref{fig:ffg-planning-1step} provides an example of how this inference process can be mapped to a factor graph, using $M_y = M_u = 2$. The node marked "MARX" is the composite node depicted in Figure \ref{fig:ffg-paramest}. It is now connected to another composite node marked "MARX-EFE", which is connected to previous observations, parameters, goal prior parameters $y_{*}$ and the to-be-taken action $u_{t}$. 
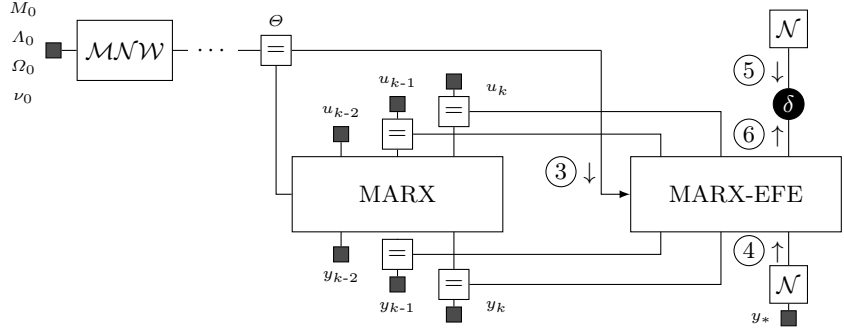
\begin{figure}[thb]
    \centering
    \scalebox{1.0}{\begin{tikzpicture}

    % t=1
    \node [draw, rectangle, minimum width=28mm, minimum height=10mm] (MARX) at (-1.0,0.0) {MARX};
    
    \node [style=observation] (uk_obs) at (-.25, 1.5) {};
    \node [style=deterministic] (eq_uk_t1) at (-.25,1.1) {$=$};
    \node [style=observation] (ukmin1_obs) at (-1.,1.2) {};
    \node [style=deterministic] (eq_ukmin1_t1) at (-1.,0.8) {$=$};
    \node [style=observation] (ukmin2_obs) at (-1.75,0.8) {};
    \node [above right=-1mm and 1mm of eq_uk_t1] (uk)  {$\scriptstyle{u_{k}}$};
    \node [above of=ukmin1_obs, node distance=3mm] (u_kmin1)  {$\scriptstyle{u_{k\text{-}1}}$};
    \node [above of=ukmin2_obs, node distance=3mm] (u_kmin2)  {$\scriptstyle{u_{k\text{-}2}}$};
    
    \draw[-] (uk_obs)   -- (eq_uk_t1);
    \draw[-] (eq_uk_t1)   -- ++(0.0, -0.6);
    \draw[-] (ukmin1_obs) -- (eq_ukmin1_t1);
    \draw[-] (eq_ukmin1_t1) -- ++(0.0, -0.3);
    \draw[-] (ukmin1_obs) -- ++(0.0, -0.1);
    \draw[-] (ukmin2_obs) -- ++(0.0, -0.3);

    \node [style=deterministic] (eq_ykmin1_t1)  at (-1.0,-0.8) {$=$};
    \node [style=deterministic] (eq_y1_t1)  at (-.25,-1.2) {$=$};
    \node [style=observation] (yk_obs) at (-.25,-1.6) {};
    \node [style=observation] (ykmin1_obs) at (-1.0,-1.2) {};
    \node [style=observation] (ykmin2_obs) at (-1.75,-0.8) {};
    % \node [right=0mm of y_kminM_node] (y_kmin_dots) {$..$};
    % \node [style=observation, below of=y_star_node,  node distance=3mm] (y1) {};

    \node [below right=-1mm and 1mm of eq_y1_t1] (y1) {$\scriptstyle{y_{k}}$};
    \node [below of=ykmin1_obs, node distance=3mm] (ykmin1) {$\scriptstyle{y_{k\text{-}1}}$};
    \node [below of=ykmin2_obs, node distance=3mm] (ykmin2) {$\scriptstyle{y_{k\text{-}2}}$};
    
    \draw[-] (eq_ykmin1_t1) -- ++(0.0, 0.2);
    \draw[-] (ykmin2_obs) -- ++(0.0, 0.3);
    \draw[-] (eq_y1_t1) -- ++(0.0, 0.7);
    \draw[-] (eq_ykmin1_t1) -- (ykmin1_obs);
    \draw[-] (yk_obs) -- (eq_y1_t1);

    % t = 2
    \node [draw, rectangle, minimum width=28mm, minimum height=10mm] (MARXEFE) at (3.5,0.0) {MARX-EFE};

    \node [style=stochastic] (u2_prior) at (4.2,2.2) {$\mathcal{N}$};
    \node [draw, circle, fill = black, text=white, inner sep=.5mm, align=center] (u2_delta) at (4.2,1.2) {$\delta$};
    \node [style=stochastic] (yt_prior) at (4.2,-1.2) {$\mathcal{N}$};

    \draw[-] (u2_prior)   -- (u2_delta);
    \draw[-] (u2_delta)   -- ++(0.0, -0.7);
    \draw[-] (eq_ukmin1_t1) -| (2.5, .5);
    \draw[-] (eq_uk_t1) -| (3.3,.5);

    \node [style=observation] (yt_obs) at (4.2,-1.65) {};
    \node [left=0mm of yt_obs] (yt) {$\scriptstyle{y_{*}}$};

    \draw[-] (eq_ykmin1_t1) -| (2.5, -.5);
    \draw[-] (eq_y1_t1) -| (3.3, -.5);
    \draw[-] (yt_prior) -- ++(0.,0.7);
    \draw[-] (yt_obs) -- (yt_prior);

    % parameter chain    
    \node [style=deterministic, above left=12mm and 0mm of MARX] (eqTheta1) {$=$};
    \node [right=40mm of eqTheta1] (eqTheta2) {};
    \node [above of=eqTheta1, node distance=4mm] {$\scriptstyle{\Theta}$};
    \node [left = 3mm of eqTheta1] (prevThetadots) {$\cdots$};
    \draw [-] (eqTheta1) |- (MARX.west);
    \draw [-] (prevThetadots) -- (eqTheta1);

    \node [shape=rectangle, draw=black, inner ysep=2mm, minimum width=12mm, minimum height=8mm, left=2mm of prevThetadots] (zetaprior) {$\mathcal{MNW}$};
    \draw [-] (zetaprior) -- (prevThetadots);
    \node [style=observation, left = 2mm of zetaprior] (zetaprior_params) {};
    \node [left of=zetaprior_params, node distance=4mm] (zetaprior_params0) {$\begin{matrix} \scriptstyle{M_0} \\ \scriptstyle{\Lambda_0} \\ \scriptstyle{\Omega_0} \\ \scriptstyle{\nu_0} \end{matrix}$};
    \draw [-] (zetaprior_params) -- (zetaprior);
    \draw [-] (eqTheta1) -- (eqTheta2.center);
    \draw[-latex] (eqTheta2.center) |- (MARXEFE.west);

    % messages
    \node [] at (1.3,.3) {$\wcircled{3} \downarrow$};
    \node [] at (3.8,-.75) {$\wcircled{4} \uparrow$};
    \node [] at (3.8,1.65) {$\wcircled{5} \downarrow$};
    \node [] at (3.8,0.8) {$\wcircled{6} \uparrow$};

\end{tikzpicture}}
    \caption{Factor graph of the 1-step ahead planning model. The left half of the graph is the same as in Figure~\ref{fig:ffg-paramest}. The parameter posterior (message $3$) is passed forwards to the MARX-EFE node, which takes in message $4$ from the goal prior node and produces message $6$ containing the exponentiated EFE function. Combined with message $5$ from the control prior node, this produces the variational control posterior. The $\delta$ circle denotes a collapse of the posterior to a Dirac delta distribution \cite{van2024realizing}.} \vspace{-5pt}
    \label{fig:ffg-planning-1step}
\end{figure}
Message $\wcircled{3}$ is the same as in Figure \ref{fig:ffg-paramest}, namely the parameter posterior distribution (Sec. \ref{sec:fe-params}). Note that during planning, the parameters are not updated. This is indicated by a \textit{directed} edge from the equality node to the MARX-EFE node. Message $\wcircled{4}$ is the goal prior distribution (Eq.~\ref{eq:control-goal-prior}), $\wcircled{5}$ is the control prior distribution (Eq.~\ref{eq:control-goal-prior}) and message $\wcircled{6}$ is the unnormalized exponentiated expected free energy;
\begin{align}
    \wcircled{4} = \mathcal{N}(y_t \given m_{*}, S_{*}) \ , \quad \wcircled{5} = \mathcal{N}(u_t \given 0, \Upsilon^{-1}) \ , \qquad \wcircled{6} = \exp(-G(u_t)) \, .
\end{align}
Note that message $6$ is the result of the EFE derivation (Thm.~\ref{eq:thm-qu}) and not the result of minimizing the Bethe free energy, a point discussed in more detail in Section \ref{sec:discussion}.

Normalizing $q^{*}(u_t)$, i.e., the product of $\wcircled{5}$ and $\wcircled{6}$, requires integrating over $u_t$. This is challenging and avoided by collapsing the approximate posterior to its maximum a posteriori point-mass distribution, $q^{*}(u_t) \approx \delta(u_t - \hat{u}_t)$ where
\begin{align} \label{eq:u_map}
    \hat{u}_t = \underset{u_t \in \mathbb{U}}{\arg \max} \ q^{*}(u_t) = \underset{u_t \in \mathbb{U}}{\arg \min} \ \frac{1}{2} u_t^{\intercal} \Upsilon u_t + G(u_t) \, .
\end{align}
We believe this is justified because collapsing the posterior to a point estimate is anyway required to pass controls to actuators.

\subsubsection{Horizon} 
\begin{figure}[thb]
    \centering
    \scalebox{1.0}{\begin{tikzpicture}

    % t=1
    \node [draw, rectangle, minimum width=18mm, minimum height=10mm] (MARX1) at (-1.0,0.0) {MARX-EFE};
    
    \node [style=stochastic] (u1_prior) at (-.25,2.7) {$\mathcal{N}$};
    \node [draw, circle, inner sep=.5mm, fill=black, text=white, align=center] (u1_delta) at (-.25,2.1) {$\delta$};
    \node [style=deterministic] (eq_u1) at (-.25,1.5) {$=$};
    \node [style=deterministic] (eq_utmin1_t1) at (-1.,1.) {$=$};
    \node [style=observation] (u_tmin1_obs) at (-1.,1.6) {};
    \node [style=observation] (u_tmin2_obs) at (-1.75,0.8) {};
    \node [above right=1mm and 0mm of eq_u1, node distance=5mm] (u1)  {$\scriptstyle{u_{t}}$};
    \node [above of=u_tmin1_obs, node distance=3mm] (u_tmin1)  {$\scriptstyle{u_{t\text{-}1}}$};
    \node [above of=u_tmin2_obs, node distance=3mm] (u_tmin2)  {$\scriptstyle{u_{t\text{-}2}}$};
    
    \draw[-] (u1_prior)     -- (u1_delta);
    \draw[-] (u1_delta)     -- (eq_u1);
    \draw[-] (eq_u1)        -- ++(0.0, -1.);
    \draw[-] (eq_utmin1_t1) -- ++(0.0, -0.5);
    \draw[-] (u_tmin1_obs)  -- (eq_utmin1_t1);
    \draw[-] (u_tmin2_obs)  -- ++(0.0, -0.3);

    \node [style=deterministic] (eq_ytmin1_t1)  at (-1.0,-1.0) {$=$};
    \node [] (eq_y1_t1)  at (-.25,-1.5) {};
    \node [style=observation] (ytmin1_obs) at (-1.0,-1.75) {};
    \node [style=observation] (ytmin2_obs) at (-1.75,-1.0) {};
    % \node [right=0mm of y_kminM_node] (y_kmin_dots) {$..$};
    % \node [style=observation, below of=y_star_node,  node distance=3mm] (y1) {};

    \node [below=-1mm of eq_y1_t1, node distance=5mm] (y1) {$\scriptstyle{y_{t}}$};
    \node [below of=ytmin1_obs, node distance=3mm] (ytmin1) {$\scriptstyle{y_{t\text{-}1}}$};
    \node [below of=ytmin2_obs, node distance=3mm] (ytmin2) {$\scriptstyle{y_{t\text{-}2}}$};
    
    \draw[-] (eq_ytmin1_t1) -- ++(0.0, 0.5);
    \draw[-] (ytmin2_obs) -- ++(0.0, 0.5);
    \draw[-] (eq_y1_t1.center) -- ++(0.0, 1.0);
    \draw[-] (eq_ytmin1_t1) -- (ytmin1_obs);

    % t = 2
    \node [draw, rectangle, minimum width=18mm, minimum height=10mm] (MARX2) at (1.8,0.0) {MARX-EFE};

    \node [style=stochastic] (u2_prior) at (2.5,2.7) {$\mathcal{N}$};
    \node [draw, circle, inner sep=.5mm, fill=black, text=white,align=center] (u2_delta) at (2.5,2.1) {$\delta$};
    \node [style=deterministic] (eq_u2) at (2.5,1.) {$=$};
    \node [style=deterministic] (eq_u1_t2) at (1.75,1.5) {$=$};
    \node [above right=5mm and -1mm of eq_u2, node distance=5mm] (u2)  {$\scriptstyle{u_{t\+1}}$};

    \draw[-] (u2_prior)      -- (u2_delta);
    \draw[-] (u2_delta)      -- (eq_u2);
    \draw[-] (eq_u2)         -- ++(0.0, -.5);
    \draw[-latex] (eq_u1)    -- (eq_u1_t2);
    \draw[-latex] (eq_u1_t2) -- ++(0.0, -1.);
    \draw[-] (eq_utmin1_t1)  -| (1., .5);
    
    \node [style=deterministic] (eq_y1_t2)  at (1.75,-1.5) {$=$};
    \node [] (eq_y2_t2)  at (2.5,-1.0) {};
    \node [below=-1mm of eq_y2_t2, node distance=5mm] (y2) {$\scriptstyle{y_{t\text{+}1}}$};

    \draw[-] (eq_ytmin1_t1) -| (1., -.5);
    \draw[-] (eq_y1_t1.center) -- (eq_y1_t2);
    \draw[-] (eq_y1_t2) -- ++(0.0, 1.0);
    \draw[-] (eq_y2_t2.center) -- ++(0.0, 0.5);

    % t = 3
    \node [draw, rectangle, minimum width=18mm, minimum height=10mm] (MARX3) at (4.6,0.0) {MARX-EFE};

    \node [style=stochastic] (u3_prior) at (5.25,2.7) {$\mathcal{N}$};
    \node [draw, circle, inner sep=.5mm,fill=black, text=white, align=center] (u3_delta) at (5.25,2.1) {$\delta$};
    \node [style=deterministic] (eq_u3) at (5.25,1.5) {$=$};
    \node [style=deterministic] (eq_u2_t3) at (4.5,1.) {$=$};
    \node [above right=.5mm and -1mm of eq_u3, node distance=5mm] (u3)  {$\scriptstyle{u_{t\+2}}$};

    \draw[-] (u3_prior)   -- (u3_delta);
    \draw[-] (u3_delta)   -- (eq_u3);
    \draw[-] (eq_u3)      -- ++(0.0, -1.);
    \draw[-latex] (eq_u1_t2)   -| (3.8,0.5);
    \draw[-latex] (eq_u2)   -- (eq_u2_t3);
    \draw[-latex] (eq_u2_t3)   -- ++(0.0, -.5);
    
    \node [style=deterministic] (eq_y2_t3)  at (4.5,-1.00) {$=$};
    \node [] (eq_y3_t3)  at (5.25,-1.5) {};
    \node [below=0mm of eq_y3_t3, node distance=5mm] (y3) {$\scriptstyle{y_{t\text{+}2}}$};

    \draw[-] (eq_y1_t2) -| (3.8,-.5);
    \draw[-] (eq_y2_t2.center) -- (eq_y2_t3);
    \draw[-] (eq_y2_t3) -- ++(0.0, 0.5);
    \draw[-] (eq_y3_t3.center) -- ++(0.0, 1.0);

    % t = 4
    \node [draw, rectangle, minimum width=18mm, minimum height=10mm] (MARXGOAL) at (7.4,0.0) {MARX-EFE};

    \node [style=stochastic] (u4_prior) at (8.2,2.7) {$\mathcal{N}$};
    \node [draw, circle, inner sep=.5mm,fill=black, text=white, align=center] (u4_delta) at (8.2,2.1) {$\delta$};
    \node [below right=4mm and -1.5mm of u4_prior, node distance=5mm] (u4)  {$\scriptstyle{u_{t\+3}}$};

    \draw[-] (u4_prior)   -- (u4_delta);
    \draw[-] (u4_delta)   -- ++(0.0, -1.6);
    \draw[-latex] (eq_u2_t3)   -| (6.6, 0.5);
    \draw[-latex] (eq_u3)   -| (7.4, 0.5);
    
    \node [style=stochastic] (ystar_node)  at (8.15,-1.25) {$\mathcal{N}$};
    \node [style=observation, below=3mm of ystar_node] (ystar_params_node) {};
    \node [right=1mm of ystar_params_node, node distance=5mm] (y_star_params) {$\scriptstyle{y_{*}}$};
    \node [above right=0mm and -1mm of ystar_node, node distance=5mm] (y4) {$\scriptstyle{y_{t\text{+}3}}$};

    \draw[-] (eq_y2_t3) -| (6.6,-.5);
    \draw[-] (eq_y3_t3.center) -| (7.4,-.5);
    \draw[-] (ystar_node) -- ++(0.0,0.75);
    \draw[-] (ystar_node) -- (ystar_params_node);

    % parameter chain
    \node [style=deterministic, above left=26mm and 0mm of MARX1] (eqTheta1) {$=$};
    \node [style=deterministic, above left=26mm and 0mm of MARX2] (eqTheta2) {$=$};
    \node [style=deterministic, above left=26mm and 0mm of MARX3] (eqTheta3) {$=$};
    \node [above of=eqTheta1, node distance=4mm] {$\scriptstyle{\Theta}$};
    \node [left = 4mm of eqTheta1] (prevThetadots) {};
    \draw [-latex] (eqTheta1) |- (MARX1.west);
    \draw [-latex] (eqTheta2) |- (MARX2.west);
    \draw [-latex] (eqTheta3) |- (MARX3.west);
    \draw [-latex] (prevThetadots) -- (eqTheta1);
    \draw [-latex] (eqTheta1) -- (eqTheta2);
    \draw [-latex] (eqTheta2) -- (eqTheta3);
    \draw [-] (eqTheta3) -- (6.1,3.31);
    \draw [-latex] (6.1,3.31) |- (MARXGOAL.west);

    % messages
    \node [] at (.5,-1.9) {$\underset{\wcircled{7}}{\rightarrow}$};
    \node [] at (1.4,-.9) {$ \wcircled{8} \! \downarrow$};
    % \msgcircle{below}{right}{eq_y1_t1}{eq_y1_t2}{0.2}{7};

\end{tikzpicture}}
    \caption{Factor graph of a $4$-step ahead planning model, showing repeated MARX-EFE node from Figure~\ref{fig:ffg-planning-1step}. Some buffer variables are now latent as well. Message $7$ is the posterior predictive over $y_t$ carrying forward system output predictions given a selection control input. Message $8$ is a predictive likelihood over $y_t$ sent backwards from the node at time $t+1$. Together the forward and backward pass of predictive messages generates a sequence of goal priors. } \vspace{-8pt}
    \label{fig:ffg-planning-Tstep}
\end{figure}
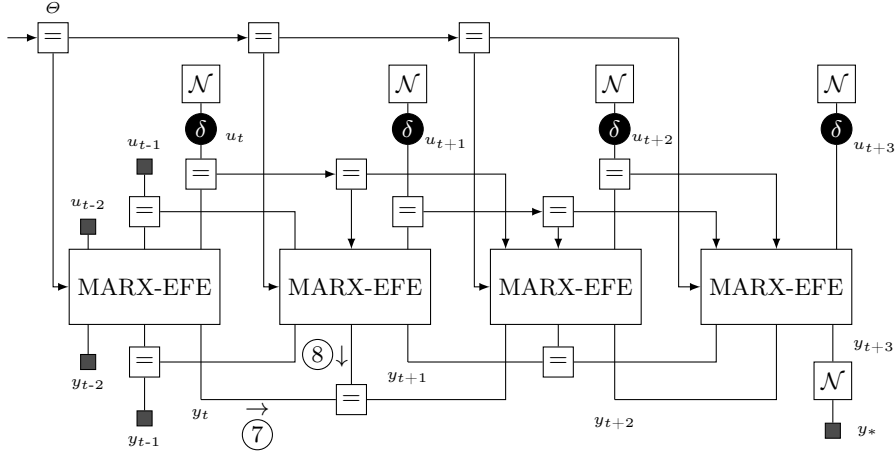

Extending the time horizon is challenging, requiring additional marginalizations that complicate the above results (see Sec.~\ref{sec:discussion} for an extended discussion). 
In this paper, we adopt a simpler approach and generalize the planning factor graph (Figure \ref{fig:ffg-planning-1step}) by including additional MARX-EFE nodes. Figure \ref{fig:ffg-planning-Tstep} shows an extension with $M_y = M_u = 2$ and a time horizon of $H = 4$.  
The main difference is that, for $t > k + 2$, the buffer $\bar{y}$ will no longer contain delta distributed variables. Note that the buffer $\bar{u}$ \textit{will} still contain delta variables, because we constrain the marginal action posteriors to be delta's (Eq.~\ref{eq:u_map}). Inspecting the second MARX-EFE node in Figure \ref{fig:ffg-planning-Tstep} reveals that the only change from the MARX-EFE node in Figure \ref{fig:ffg-planning-1step} is the incoming message for $y_t$. This message is the posterior predictive distribution (given a selected action $\hat{u}_t$) sent out by the first MARX-EFE node in the planning graph; 
\begin{align}
    \wcircled{7} = p(y_t \given \hat{u}_t, \mathcal{D}_t) = \mathcal{T}_{\eta_k}(y_t \given \mu_k(\hat{u}_t), \Sigma_k(\hat{u}_t)) \, .
\end{align} 
This message is incorporated into the MARX-EFE node function through a variational approximation:
\begin{align}
    p(&y_{t\+1} \given u_{t\+1}, \bar{u}_{t\+1}, \tilde{y}_{t\+1}, \Theta) \nonumber \\
    &\ \approx \exp\big( \mathbb{E}_{ p(y_t | \hat{u}_t, \mathcal{D}_t) }\big[ \ln p(y_{t\+1} \given u_{t\+1}, \bar{u}_{t\+1}, \bar{y}_{t\+1}, \Theta) \big] \big) \label{eq:variationalapprox1} \\
    &\ \propto \exp\big( - \frac{1}{2} (y_{t\+1}^{\intercal} W y_{t\+1} \! - \! 2 y_{t\+1} W A^{\intercal}  \mathbb{E}_{p(y_t | \hat{u}_t, \mathcal{D}_t)} \begin{bmatrix} u_{t\+1} & \bar{u}_{t\+1} & y_t & \hat{y}_{t\-1}  \end{bmatrix}^{\intercal} ) \big) \\
    &\ \propto \mathcal{N}(y_{t\+1} \given A^{\intercal} \begin{bmatrix} u_{t\+1} & \bar{u}_{t\+1} & \mu_k(\hat{u}_t) & \hat{y}_{t\-1}  \end{bmatrix}^{\intercal}, W^{-1}) .
\end{align}
Note that this is, in essence, still the MARX likelihood function except with the mean of the posterior predictive $\mu_k(\hat{u}_t)$ instead of an observed value for $y_t$ in $\bar{y}_{t\+1}$. We mark this change with $\tilde{y}$ instead of $\bar{y}$.

For the backwards message from the MARX-EFE node at $t+1$ towards the variable $y_t$, we first utilize the same variational approximation as above but now with respect to the variational factor $q(y_{t\+1}) = \mathcal{N}(y_{t\+1} \given m_{t\+1}, S_{t\+1})$ (Eq.~\ref{eq:qyt});
 \begin{align}
    \exp\big( &\mathbb{E}_{q(y_{t\+1})}\big[ \ln p(y_{t\+1} \given u_{t\+1}, \bar{u}_{t\+1}, \bar{y}_{t\+1}, \Theta) \big] \big) \nonumber \\
    &\propto \exp\big( -\frac{1}{2} \mathbb{E}_{q(y_{t\+1})}\big[y_{t\+1}^{\intercal} W y_{t\+1} - 2 y_{t\+1} W A^{\intercal}   \begin{bmatrix} u_{t\+1} & \bar{u}_{t\+1} & y_t & \hat{y}_{t\-1}  \end{bmatrix}^{\intercal} \big] \big) \label{eq:variationalapprox2} \\
    &\propto \mathcal{N}(m_{t\+1} \given A^{\intercal} \begin{bmatrix} u_{t\+1} & \bar{u}_{t\+1} & y_t & \hat{y}_{t\-1}  \end{bmatrix}^{\intercal}, W^{-1}) .
\end{align}
Then we marginalize over the parameter posterior distribution,
\begin{align}
    \wcircled{8} &\propto \mathbb{E}_{p(\Theta \given \mathcal{D}_k)}\big[ \, \mathcal{N}(m_{t\+1} \given A^{\intercal} \begin{bmatrix} u_{t\+1} & \bar{u}_{t\+1} & y_t & \hat{y}_{t\-1}  \end{bmatrix}^{\intercal}, W^{-1}) \big] \\
    &\propto \mathcal{T}_{\bar{\eta}}\big(m_{t\+1} \given \bar{\mu}_{t\+1}(y_t), \bar{\Sigma}_{t\+1}(y_t) \big) \, ,
\end{align}
with $\bar{\eta} = \nu_k - D_y + 1$ degrees of freedom and mean and covariance
\begin{align}
    \bar{\mu}_{t\+1}(y_t) = M_k^{\intercal} \begin{bmatrix} u_{t\+1} \\ \bar{u}_{t\+1} \\ y_t \\ \hat{y}_{t\-1}  \end{bmatrix}  , \ \
    \bar{\Sigma}_{t\+1}(y_t) = \frac{1}{\bar{\eta}} \Omega_k \Big(1  +  \begin{bmatrix} u_{t\+1} \\ \bar{u}_{t\+1} \\ y_t \\ \hat{y}_{t\-1}  \end{bmatrix}^{\intercal} \Lambda_k^{-1} \begin{bmatrix} u_{t\+1} \\ \bar{u}_{t\+1} \\ y_t \\ \hat{y}_{t\-1}  \end{bmatrix} \Big) \, .
\end{align}
In essence, this distribution scores which values of $y_t$ best predict $y_{t\+1}$, with $m_{t\+1}$ as a pseudo-observation.
At the $y_t$ edge, we perform a variational factor update based on the product of messages $\wcircled{7}$ and $\wcircled{8}$:
\begin{align}
    q(y_t) &\propto \mathcal{T}_{\eta_{k}}\big(y_{t} \given \mu_{k}(\hat{u}_t), \Sigma_{k}(\hat{u}_t) \big)  \mathcal{T}_{\bar{\eta}}\big(m_{t\+1} \given \bar{\mu}_{t\+1}(y_t), \bar{\Sigma}_{t\+1}(y_t) \big) \label{eq:qyt} \, .
\end{align}
This product is not part of a known parametric family of distributions. We perform a Laplace approximation to produce $q(y_t) \approx \mathcal{N}(y_t \given m_t, S_t)$ where \cite{mackay2003information}:
\begin{align} \label{eq:laplace-approx}
    m_t &= \underset{y_t}{\arg \max} \ \ln \mathcal{T}_{\eta_k}\big(y_{t} \given \mu_{k}(\hat{u}_t), \Sigma_{k}(\hat{u}_t) \big) \mathcal{T}_{\bar{\eta}}\big(m_{t\+1} \given \bar{\mu}_{t\+1}(y_t), \bar{\Sigma}_{t\+1}(y_t) \big) \\
    S_t^{-1} \! &= \! - \nabla^2_{y_t} \ln \mathcal{T}_{\eta_k}\big(y_{t} | \mu_{k}(\hat{u}_t), \Sigma_{k}(\hat{u}_t) \big) \mathcal{T}_{\bar{\eta}}\big(m_{t\+1} | \bar{\mu}_{t\+1}(y_t), \bar{\Sigma}_{t\+1}(y_t) \big) \Big|_{y_t = m_t} .
\end{align}
This Gaussian variational factor effectively becomes a goal prior for time $t$. So, in the extended time horizon, we see that the forward and backward passes over the future observations generate a sequence of intermediate goal priors. As such, at each future time point, the agent needs only to solve a 1-step ahead expected free energy minimization problem. 

\section{Experiments}
We perform simulation experiments in which agents have to reach a target state in a single trial. We refer to the proposed free energy minimizing agent as \emph{MARX-EFE}\footnote{Agent built with RxInfer; \url{https://github.com/biaslab/IWAI2025-MARXEFE-MP}}. The benchmark is the same agent but with controls found by minimizing a standard model-predictive control cost function;
\begin{align}
    \hat{u}^{\text{MPC}}_{k+1:k+H} = \underset{u_{k+1:k+H} \, \in \, \mathbb{U}^H}{\arg \min} \ \ \sum_{t=k+1}^{k+H} u_t^{\intercal}\Upsilon u_t  + \big(\mu_t(u_t)  -  m_{*} \big)^{\intercal} \big(\mu_t(u_t)  - m_{*} \big) .
\end{align} 
This agent will be called \emph{MARX-MPC}.
We evaluate the probabilities of the system output $y_k$ under the goal prior distribution, $p(y_k \given y_{*})$. Additionally, we evaluate model evidence, i.e., the probability of the system output observation under the agent's predictive distribution, $p(y_k \given u_k, \mathcal{D}_k)$. 

\subsubsection{System} \label{sec:experiments:robotnav}
Consider a linear Gaussian dynamical system where the state vector $z_k$ contains the two-dimensional position and velocity of a robot. The robot's state transition and measurement functions are
\begin{align}
    f(z_{k\-1}, u_k) &= \begin{bmatrix} 1 & 0 & \Delta t & 0 \\ 0 & 1 & 0 & \Delta t \\ 0 & 0 & 1 & 0 \\ 0 & 0 & 0 & 1 \end{bmatrix} z_{k\-1} + \begin{bmatrix} 0 & 0 \\ 0 & 0 \\ \Delta t & 0 \\ 0 & \Delta t \end{bmatrix} u_k \, , \quad g(z_k) = \begin{bmatrix} 1 & 0 & 0 & 0 \\ 0 & 1 & 0 & 0 \end{bmatrix} z_k ,
\end{align}
where $\Delta t$ is the time step size. Its covariance matrices are
\begin{align}
     Q &= \begin{bmatrix} \frac{\Delta t^3}{3} \varsigma_1 & 0 & \frac{\Delta t^2}{2} \varsigma_1 & 0 \\
                      0 & \frac{\Delta t^3}{3} \varsigma_2 & 0 & \frac{\Delta t^2}{2} \varsigma_2 \\
                      \frac{\Delta t^2}{2} \varsigma_1 & 0 & \Delta t \varsigma_1 & 0 \\
                      0 & \frac{\Delta t^2}{2} \varsigma_2 & 0 & \Delta t \varsigma_2 \end{bmatrix} \, , \quad R = \begin{bmatrix}  \rho_1 & 0 \\ 0 & \rho_2 \end{bmatrix} \, ,
\end{align}
with $\varsigma = \begin{bmatrix} 10^{-6} \ 10^{-6} \end{bmatrix}^{\intercal}$ and $\rho = \begin{bmatrix} 10^{-3} \ 10^{-3} \end{bmatrix}^{\intercal}$.

\subsubsection{Prior parameters} \label{sec:experiment-priors} 
The prior parameters are weakly informative; $\nu_0 = 100$, $M_0 = 1/(D_x D_y) \cdot I_{D_x \times D_y}$, $\Lambda_0 = 10^{-2} \cdot I_{D_x}$, $\Omega_0 = I_{D_y}$, and $\Upsilon = 10^{-6} \cdot I_{D_u}$. The system starts at $z_0 = \begin{bmatrix} 0 & 0 & 0 & 0 \end{bmatrix}$ and the goal prior has mean $m_* = \begin{bmatrix} 0 & 1 \end{bmatrix}^{\intercal}$ and covariance matrix $S_{*} = 10^{-6} \cdot I_{D_y}$. Buffers are fixed at $M_u = M_y = 2$ and the time horizon at $H = 3$. Controls are limited to $\mathbb{U} = [-1, \, 1]$ for $T=10 000$ steps at $\Delta t = 0.1$.

\subsubsection{Results}
\begin{figure*}[thb]
    \centering
    \includegraphics[height=85pt]{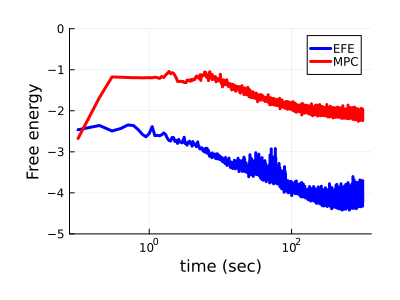}
    \includegraphics[height=85pt]{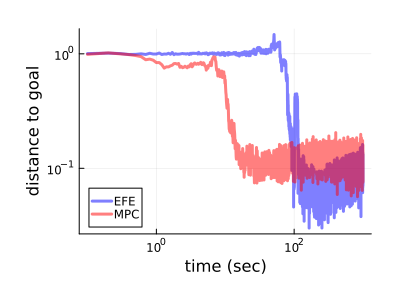}
    \includegraphics[height=85pt]{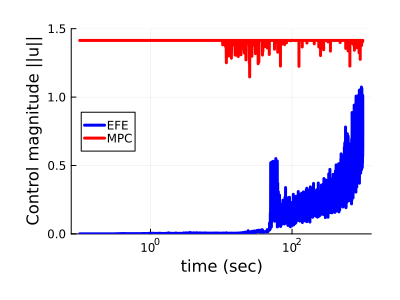}
    \caption{MARX-EFE (blue) vs. MARX-MPC (red) over a trial of $1000$ seconds, compared in terms of free energy (left), Euclidean distance to goal ($m_{*}$; middle) and 2-norm of controls (right). Results are averaged over 10 experiments.  MARX-EFE initially takes smaller actions, aiming to improve parameters and predictions first. It arrives at the goal later than MARX-MPC but is better able to park on the goal itself. MARX-EFE's actions are small initially but increase in magnitude as uncertainty shrinks.} 
    \label{fig:method-comparison}
\end{figure*}

Figure \ref{fig:method-comparison} shows the experimental results comparing MARX-EFE to MARX-MPC. The left figure shows that MARX-EFE consistently scores a smaller free energy than MARX-MPC, demonstrating that it cares more strongly about accurately predicting its next observation. The middle figure shows the distance to goal over the duration of the trial where, on average, MARX-MPC reaches the goal sooner than MARX-EFE. MARX-MPC does not care about making accurate predictions, only to close the gap to the target as quickly as possible. It is successful in that regard but struggles to park itself on the target exactly because it ignored opportunities to learn the finer parts of robot's dynamics earlier in the trial. The MARX-EFE agent ultimately gets closer than MARX-MPC because - by the time it gets to the goal - it has a much better model of the robot's dynamics. The right figure shows the 2-norm of the controls, highlighting that MARX-MPC consistently utilizes maximum power ($\max \|u_t\|_2 = \sqrt{2}$) to get closer. The MARX-EFE agent takes very small actions in the beginning, when it is uncertain of their outcomes, and slowly takes larger actions when its uncertainty shrinks. We interpret this behaviour as some form of "caution". 

\section{Discussion} \label{sec:discussion}

For planning, we observe that the expectation over the future observation is actually not necessary in the above model. If the likelihood is incorporated into the numerator of Eq.~\ref{eq:EFE-planning} as well, i.e.,
\begin{align} \label{eq:FE-planning}
\mathcal{F}_k[q] = \mathbb{E}_{q(y_t, \Theta, u_t)} \Big[ \ln \frac{q(y_t, \Theta, u_t)}{p(y_t, \Theta, u_t \given y_{*}, \mathcal{D}_k)} \Big] \, ,
\end{align}
then -  following the same steps as in Appendix \ref{app:proofs} - the EFE function becomes:
\begin{align}
    G(u_t) = \mathbb{E}_{p(y_t \given u_t, \mathcal{D}_k)} \big[ \ln  p(y_t \given u_t, \mathcal{D}_k)  \big] + \mathbb{E}_{p(y_t \given u_t, \mathcal{D}_k)} \big[ -\ln p(y_t \given y_{*}) \big] \, .
\end{align} 
In both the mutual information in Eq.~\ref{eq:EFEfunction} and in the entropy of the posterior predictive above, the only term that depends on $u_t$ is the variance of the posterior predictive $\Sigma_k(u_t)$. All other terms drop out due to the translation invariance of differential entropies. Thus, we find the same solution when using a standard free energy functional instead of an expected free energy functional \cite{de2025expected}.

\paragraph{Limitations} In Sec.~\ref{sec:fe-actions} we avoided forming the joint posterior predictive distribution over all future outputs in the horizon;
\begin{align}
    p(y_{k+1:k+H} &\given u_{k+1:k+H}, \mathcal{D}_k) = \int p(\Theta \given \mathcal{D}_k)  \prod_{t=k+1}^{k+H} \, p(y_t \given \Theta, u_t, \bar{u}_t, \bar{y}_t) \ \mathrm{d}\Theta \, \, . 
\end{align}
This marginalization is a challenge because blocks of the autoregressive coefficient start to nest in both the mean and covariance matrix of the joint Gaussian likelihood. To illustrate this, consider an example with $M_y = 1$ such that $\bar{y}_t = y_{t\-1}$. The joint distribution of the likelihoods for $k+1$ and $k+2$ is Gaussian distributed with mean vector and covariance matrix
\begin{align}
    \begin{bmatrix} A_1^{\intercal} u_{k\+1} + A_2^{\intercal} \bar{u}_{k} +  A_3^{\intercal} y_k \\ A_1^{\intercal} u_{k\+2} \! + \! A_2^{\intercal} \bar{u}_{k\+1} \! + \! A_3^{\intercal} \big(A_1^{\intercal} u_{k\+1} \! + \! A_2^{\intercal} \bar{u}_{k} \! + \! A_3^{\intercal} y_k \big)  \end{bmatrix} \begin{bmatrix} W^{\-1} & W^{\-1} A_3 \\ A_3^{\intercal} W^{\-1} & A_3^{\intercal} W^{\-1} A_3  +  W^{\-1} \end{bmatrix} ,
\end{align}
where $A_{i}$ represent row-indexed blocks of the coefficient matrix $A$. Marginalizing this joint future likelihood over $p(\Theta \given \mathcal{D}_k)$ is difficult but a solution would avoid the variational and Laplace approximation errors in Eqs.~\ref{eq:variationalapprox1}, \ref{eq:variationalapprox2}, and \ref{eq:laplace-approx}.

\section{Conclusion}
We designed an active inference agent with continuous-valued actions as a message passing procedure on a factor graph. The forward and backward pass of predictions over future system outputs generates a sequence of intermediate goal priors. Each node in the planning graph only has to solve a 1-step ahead EFE minimization problem to find appropriate controls. The agent successfully navigates a robot to a goal position under unknown dynamics.

\begin{credits}
\subsubsection{\ackname} The authors gratefully acknowledge support from the Eindhoven Artificial Intelligence Systems Institute.

\subsubsection{\discintname}
The authors have no competing interests to declare that are relevant to the content of this article.
\end{credits}

\appendix
\section{Appendix} \label{app:proofs}

% \paragraph{Theorem 1}
\begin{proof} \label{proof:thm1}
Using the factorisation of the variational model, the expected free energy functional can be re-arranged to isolate the expectation over $q(u_t)$:
\begin{align}
    \mathcal{F}_t[q] &= \mathbb{E}_{q(y_t \given \Theta, u_t, \bar{u}_t, \bar{y}_t)} \Big[ \mathbb{E}_{q(\Theta,u_t)} \big[ \ln \frac{q(\Theta,u_t)}{p(y_t, \Theta, u_t \given y_{*}, \mathcal{D}_k)} \big] \Big]  \\
    &= \mathbb{E}_{q(u_t)} \Big[ \ln \frac{q(u_t)}{p(u_t)} + \mathbb{E}_{q(y_t \given \Theta, u_t, \bar{u}_t, \bar{y}_t) q(\Theta)} \big[ \ln \frac{q(\Theta)}{p(y_t, \Theta \given u_t, y_{*}, \mathcal{D}_k)} \big] \Big] \\
    &= \mathbb{E}_{q(u_t)} \Big( \ln \frac{q(u_t)}{p(u_t) \exp(-G(u_t))} \Big) \, ,
\end{align}
for $G(u_t) = \mathbb{E}_{q(y_t \given \Theta, u_t, \bar{u}_t, \bar{y}_t) q(\Theta)} \big[ \ln q(\Theta) / p(y_t, \Theta \given u_t, y_{*}, \mathcal{D}_k) \big]$ and the identity $G(u_t) = \ln \big(1/\exp(-G(u_t)) \big)$.
Constraining $q(u_t)$ to be a probability distribution over the space of affordable controls $\mathbb{U}$ is done with a Lagrange multiplier:
\begin{align}
    \mathcal{L}[q, \gamma] &= \mathcal{F}_k[q] + \gamma \Big(\int_{\mathbb{U}} q(u_t) \mathrm{d}u_t - 1 \Big) \, .
\end{align}
The stationary solution $q^{*}(u_t)$ of the Lagrangian is found at $\delta \mathcal{L}[q, \gamma] / \delta q = 0$ \cite{wainwright2008graphical}). Let $\delta q(u_t) = \epsilon \phi(u_t)$ be a variation with $\phi$ a continuous and differentiable test function. Then the variational derivative can be found with:
\begin{align}
    \int_{\mathbb{U}} \frac{\delta \mathcal{L}[q,\gamma]}{\delta q} \phi(u_t) \mathrm{d}u_t &= \frac{d \mathcal{L}[q(u_t) \! + \! \epsilon \phi(u_t),\gamma]}{d \epsilon} \Big|_{\epsilon = 0} \\
    % &= \frac{d}{d \epsilon} \mathcal{F}[q(u_t) + \epsilon \phi(u_t)] \Big|_{\epsilon = 0} + \frac{d}{d\epsilon} \gamma \big( \int_{\mathbb{U}} q(u_t) + \epsilon \phi(u_t) \mathrm{d}u_t - 1 \big) \Big|_{\epsilon = 0} \\
    % &= \frac{d}{d \epsilon} \int_{\mathbb{U}}  \big(q(u_t) \+ \epsilon \phi(u_t) \big) \ln \frac{q(u_t) + \epsilon \phi(u_t)}{p(u_t) \exp(-G(u_t))} \mathrm{d}u_t  \! + \! \gamma \int_{\mathbb{U}} (q(u_t) \+ \epsilon \phi(u_t) )\mathrm{d}u_t \Big|_{\epsilon = 0} \\
    % &= \int_{\mathbb{U}} \big( \phi(u_t) \ln \frac{q(u_t)}{p(u_t) \exp(-G(u_t))} + \phi(u_t)  \big) \mathrm{d}u_t + \gamma \int_{\mathbb{U}}  \phi(u_t) \mathrm{d}u_t \\
    &= \int_{\mathbb{U}} \Big(\ln \frac{q(u_t)}{p(u_t) \exp(-G(u_t))} + 1 + \gamma \Big) \phi(u_t)  \mathrm{d}u_t \, .
\end{align}
Setting the variational derivative to $0$, yields
\begin{align}
    % &\frac{\delta \mathcal{L}[q(u_t), \gamma]}{\delta q(u_t)} = 
    \ln \frac{q(u_t)}{p(u_t) \exp(\! -  G(u_t))} \! + 1 + \gamma = 0  \rightarrow 
    q(u_t) \! = \! \exp( \! -  \gamma \! - \! 1) p(u_t) \exp(\! -  G(u_t)) . \label{eq:funcderiv0}
\end{align}
Plugging this into the constraint gives $\exp( - \gamma - 1) = 1/\int_{\mathbb{U}} p(u_t) \exp(-G(u_t)) \mathrm{d}u_t$. As such, we have:
\begin{align}
    q(u_t) = \frac{p(u_t) \exp(-G(u_t))}{\int_{\mathbb{U}} p(u_t) \exp(-G(u_t)) \mathrm{d}u_t} \, .\label{eq:optimalq-norm}
\end{align}
Using \eqref{eq:p-future}, \eqref{eq:decomp-futurejoint} and \eqref{eq:variational-model}, $G(u_t)$ can be simplified to a negative mutual information plus a cross-entropy term:
\begin{align}
    G&(u_t) 
    % &= \mathbb{E}_{q(y_t, \Theta \given u_t) } \big[ \ln \frac{q(y_t, \Theta \given u_t)}{ p(y_t, \Theta \given u_t , y_{*}, \mathcal{D}_k) } \big] \\
    = \mathbb{E}_{p(y_t \given \Theta, u_t, \bar{u}_t, \bar{y}_t) p(\Theta \given \mathcal{D}_k)} \big[ \ln \frac{p(\Theta \given \mathcal{D}_k) p(y_t \given u_t, \mathcal{D}_k)}{p(y_t \given \Theta, u_t, \bar{u}_t, \bar{y}_t) p(\Theta \given \mathcal{D}_k) p(y_t \given y_{*})} \big] \\
    &= - \mathbb{E}_{p(y_t, \Theta | u_t, \mathcal{D}_k)} \big[ \ln  \frac{p(y_t, \Theta \given u_t, \mathcal{D}_k)}{p(y_t | u_t, \mathcal{D}_k)p(\Theta | \mathcal{D}_k)} \big] \! + \! \mathbb{E}_{p(y_t | u_t, \mathcal{D}_k)} \big[\! -\ln p(y_t | y_{*}) \big] . \label{eq:Ju-KL}
    % &= D_{\text{KL}} \big[ p(y_t \given u_t, \mathcal{D}_k) \|  p(y_t \given y_{*}) \big] \, .
\end{align}
\end{proof}

% \paragraph{Corollary 1}
\begin{proof} \label{proof:cor1}
We split the mutual information into a joint entropy minus the entropy of the posterior predictive and that of the parameter posterior \cite{mackay2003information}:
\begin{align}
-&\mathbb{E}_{p(y_t, \Theta | u_t, \mathcal{D}_k)} \big[ \ln  \frac{p(y_t, \Theta \given u_t, \mathcal{D}_k)}{p(y_t | u_t, \mathcal{D}_k)p(\Theta | \mathcal{D}_k)} \big] =  \mathbb{E}_{p(y_t \given u_t, \mathcal{D}_k)} \big[\ln p(y_t \given u_t, \mathcal{D}_k) \, \big] \nonumber \\
&\qquad  +  \mathbb{E}_{p(\Theta \given \mathcal{D}_k)} \big[ \ln p(\Theta \given \mathcal{D}_k) \, \big] - \mathbb{E}_{p(y_t, \Theta | u_t, \mathcal{D}_k)} \big[ \ln p(y_t, \Theta \given u_t, \mathcal{D}_k)  \big] \, . 
\end{align}
Since entropies are invariant to translation, only the entropy of the posterior predictive affects $G(u_t)$ \cite{kouw2023information}. The entropy of a location-scale T-distribution is \cite{kotz2004multivariate}:
\begin{align}
    &\mathbb{E}_{p(y_t | u_t, \mathcal{D}_k)} \big[ \ln  p(y_t \given u_t, \mathcal{D}_k)  \big] 
    \nonumber \\
    &= - \mathbb{E}_{\mathcal{T}_{\eta_k}(y_t \given 0, \Sigma_k(u_t))} \big[-\ln \mathcal{T}_{\eta_k}(y_t \given 0, \Sigma_k(u_t)) \big]  \\
    &= \! - \! \ln\frac{(\eta_k \pi)^{\frac{D_y}{2}}}{\Gamma(\frac{D_y}{2})}B\big(\frac{D_y}{2}, \! \frac{\eta_k}{2} \big) \! - \! \frac{\eta_k \+ D_y}{2}\big(\psi(\frac{\eta_k \! + \! D_y}{2}) \! - \! \psi (\frac{\eta_k}{2}) \big) \! - \! \frac{1}{2}\ln \big| \Sigma_k(u_t) \big| .
    % &= \text{constants} - \frac{1}{2}\ln \big| \Sigma_t(u_t) \big| \, .
\end{align}
where $B(\cdot), \Gamma(\cdot), \psi(\cdot)$ are the beta, gamma and digamma functions, respectively. Note that only the last term depends on $u_t$. The cross-entropy from posterior predictive to goal distribution is:
\begin{align}
    &\mathbb{E}_{p(y_t \given u_t, \mathcal{D}_k)} \big[ - \ln p(y_t \given y_{*}) \big] \nonumber \\
    &= \frac{1}{2} \Big( \ln 2\pi |S_*| + \mathbb{E}_{\mathcal{T}_{\eta_t}(y_t \given \mu_t(u_t), \Sigma_t(u_t))} \big[(y_t - m_{*})^{\intercal} S_*^{-1} (y_t - m_{*}) \big] \Big) \\
    % &= - \frac{1}{2} \Big( \ln 2\pi + \ln |S_*| + \tr\Big[ S_{*}^{-1} \mathbb{E}_{\mathcal{T}(y_t \given \mu_t(u_t), \Sigma_t(u_t), \eta_t)} \big[y_t y_t^{\intercal} - 2 y_t m_{*}^{\intercal} + m_{*} m_{*}^{\intercal} \big] \Big] \Big) \\
    % &= \frac{1}{2} \ln 2\pi |S_*|  + \frac{1}{2} (\mu_t(u_t) \! - \! m_{*})^{\intercal} S_*^{-1} (\mu_t(u_t) \! - \! m_{*}) + \frac{1}{2}\tr\big[S_{*}^{-1} \frac{\eta_t}{\eta_t \- 2} \Sigma_t(u_t) \big]   .
    &= \frac{1}{2} \ln 2\pi |S_*|  +  \frac{1}{2}\tr\Big[S_{*}^{-1} \big(\Sigma_t(u_t) \frac{\eta_t}{\eta_t \! - \! 2} + \Xi(u_t) \big) \Big]  ,
    % &= \text{constants}  + \frac{1}{2} \tr\Big[S_{*}^{-1} \big(\frac{\eta_t}{\eta_t \- 2}\Sigma_t(u_t) + (\mu_t(u_t) \! - \! m_{*})(\mu_t(u_t) \! - \! m_{*})^{\intercal} \big) \Big]  .
\end{align}
where $\Xi(u_t) = (\mu_t(u_t) \! - \! m_{*})(\mu_t(u_t) \! - \! m_{*})^{\intercal}$. Note that the first term is constant.
\end{proof}

\bibliographystyle{splncs04}
\bibliography{references}

\end{document}